\documentclass{article}
\usepackage{spconf,amsmath,graphicx}

\usepackage{enumitem}
\setlist{nosep, leftmargin=14pt}

\usepackage{mwe} % to get dummy images
\usepackage{comment}
\usepackage{soul}

\usepackage[vmargin=20mm,hmargin=20mm]{geometry}  % adjust page margins
\usepackage{graphicx} % allows inclusion of PDF, PNG and JPG images
\usepackage{parskip}  % separate paragraphs with vertical space
\usepackage{setspace} % for \onehalfspacing
\usepackage{refcount} % for counting pages
\usepackage{upquote}  % for correct quotation marks in verbatim text
\usepackage{amsmath}
\usepackage{amssymb}
\usepackage{color,soul}
\usepackage{enumitem}
\usepackage{tabularray}
\usepackage{array}
\usepackage{rotating}
\usepackage{multirow}
\usepackage{pifont}
\usepackage{titlesec}
\usepackage{xcolor}

\def\x{{\mathbf x}}

\newcommand{\unfair}[1]{\textcolor{red}{#1}}

\title{Machine Learning Fairness for Depression Detection using EEG Data}
\name{Angus Man Ho Kwok, Jiaee Cheong, Sinan Kalkan, Hatice Gunes
}
\name{Angus Man Ho Kwok$^{\star}$ \qquad 
Jiaee Cheong$^{\star \dagger \ddagger}$ \qquad 
Sinan Kalkan$^{\dagger}$ \qquad 
Hatice Gunes$^{\star}$}
\address{$^{\star}$University of Cambridge %\\
  $^{\dagger}$Middle East Technical University
  $^{\ddagger}$The Alan Turing Institute
  }
\begin{document}
%\ninept
%
\maketitle
\begin{abstract}

This paper presents the very first attempt to evaluate 
machine learning fairness for depression detection using electroencephalogram (EEG) data.
We conduct experiments using different deep learning architectures such as  Convolutional Neural Networks (CNN), Long Short-Term Memory (LSTM) networks, and Gated Recurrent Unit (GRU) networks across
three EEG datasets: Mumtaz, MODMA and Rest.
We employ five different bias mitigation strategies at the pre-, in- and post-processing stages and evaluate their effectiveness. 
Our experimental results show that bias exists in existing EEG datasets and algorithms for depression detection, 
and different bias mitigation methods address bias at different levels across different fairness measures.

\end{abstract}
\begin{keywords}
EEG, ML fairness, Depression Detection
\end{keywords}
%

%%%%%%%%%%%%%%%%%%%%%%%%%%%%%%%%%%%%%
%%%%%%%%%%%%%%%%%%%%%%%%%%%%%%%%%%%%%

%\begin{comment}

%%%%%%%%%%%%%%%%%%%%%%%%%%%%
%%%%%%%%%%%%%%%%%%%%%%%%%%%%

\section{Introduction}
\label{sect:intro}

Major depressive disorders (MDD) are becoming increasingly prevalent worldwide. 
Machine learning (ML), especially deep learning (DL) based methods, have been recently used in many research studies for depression detection with success \cite{yang2020decoding,zhou2018epileptic,chen2024mgsn}.
%,abidi2024eegdepressionnet,chen2024mgsn}.
%
In concurrence, ML bias is becoming a growing source of concern \cite{cheong2021hitchhiker}. 
Bias can be understood as discrimination against individuals based on certain sensitive attributes such as age, race and gender \cite{hort2022bias,pessach2022review,cheong2023counterfactual}.
Fairness conversely dictates that no individual or subgroup should be advantaged or disadvantaged based on their inherent characteristics.
Given the high stakes involved in MHD analysis, it is crucial to investigate and mitigate the ML biases present. 
Research indicated the presence of ML bias across a variety of tasks ranging from
automated video interviews \cite{booth2021bias} to image search \cite{feng2022has}.
However, none of the existing works have addressed ML fairness in MDD detection
using EEG data.

\textbf{Our contributions} in this paper are as follows. 
First, our study is the first attempt to evaluate ML fairness for depression detection using electroencephalogram (EEG) data. 
None of the existing work on ML fairness for depression detection \cite{jiaee, aguirre, postpartum, cheong_Ufair}
has investigated bias mitigation for depression detection using EEG data. 
Second, we study and compare the effectiveness of a diverse set of bias mitigation techniques
to improve fairness in EEG-based depression detection. We show they have different effects on performance and fairness. 
We conduct our experiments using three different deep architectures across three datasets, Mumtaz, MODMA and Rest.
Throughout our experimentation, we attempt to address the following two research questions (RQs). 
\textbf{RQ 1:} Is there bias within existing EEG data and ML algorithm for depression detection?
\textbf{RQ 2:} How effective are existing bias mitigation methods at improving ML fairness for depression detection using EEG data?
Our experimental results indicate that both data and algorithmic biases
exist and that different bias mitigation provides different degree of effectiveness across different datasets and algorithms.
%

%\begin{comment}

\begin{figure}[h]
    \centering
    \includegraphics[width=0.99\columnwidth]{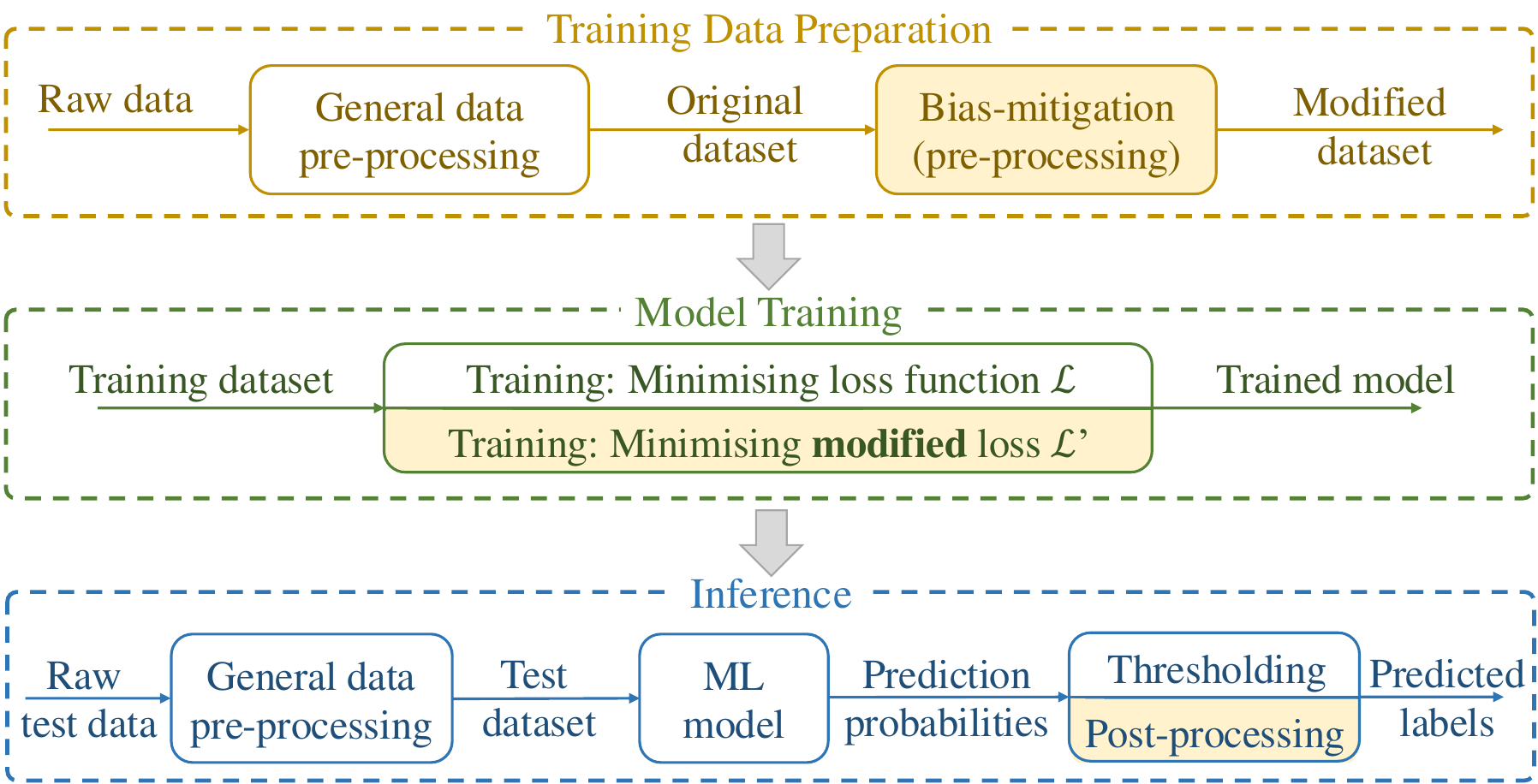}
   \emph
    {\caption{Simplified pipeline: highlighted parts indicate how bias is mitigated at the pre-, in- and post-processing stages.}
    \vspace{-5mm}
    \label{fig:pipeline}}
\end{figure}

%\end{comment}

%%%%%%%%%%%%%%%%%%%%%%%%%%%%
%%%%%%%%%%%%%%%%%%%%%%%%%%%%

\section{Methodology}
\vspace{-3mm}

We approach MDD detection as a classification problem where we have a dataset $D = \{(\mathbf{x}_i, y_i)\}_i$  where $\mathbf{x}_i\in X$ is a tensor representing information (e.g. EEG data) about an individual $I$ and $y_i\in Y$ is the outcome (e.g. 1 for depressed vs. 0 for non-depressed) that we wish to predict. Each input $\mathbf{x}_i$ is associated (through an individual $I$) with a sensitive attribute $s(\mathbf{x}_i) \in S$ where $S = \{male, female\}$. This is a classification problem where we are interested in finding a parameterised function $f$ with $f: X\rightarrow Y$. % The function 
$f(\ \cdot\ ; \theta)$ estimates the probabilities for all outcomes (classes)  $p(Y | \mathbf{x}_i)$. We use $p(y_i | \mathbf{x}_i)$ to denote the predicted probability for the correct class.
The goal of bias mitigation is to ensure that the outcomes for each demographic subgroup adhere to the different fairness measures outlined in Section \ref{sect: measures}.

%%%%%%%%%%%%%%%%%%%%%%%%%%%%
%%%%%%%%%%%%%%%%%%%%%%%%%%%%
\subsection{Bias Mitigation Methods}
\label{sect: mitigation}

We employ two pre-processing, two in-processing and one post-processing bias mitigation methods.
Each method has its own advantages and disadvantages. Pre- and post-processing may be easier to implement but in-processing may be the most effective \cite{cheong2021hitchhiker}. 
We have chosen the commonly used methods according to \cite{hort2022bias}.

\vspace{-2mm}
\paragraph{(a) Pre-processing: Data Augmentation}
We employ a data augmentation technique, 
\emph{Mixup} \cite{mixup},
which works by generating new samples for the minority group so that the resulting dataset is balanced across gender.
A new sample $(\mathbf{x}', y')$ is generated from two randomly drawn examples $(\mathbf{x}_i, y_i)$ and $(\mathbf{x}_j, y_j)$ with $i \neq j$ via:
\begin{gather}\footnotesize
    \mathbf{x}' = \lambda \mathbf{x}_i + (1 - \lambda)\mathbf{x}_j, \\
    y' = \lambda y_i + (1 - \lambda)y_j,
\end{gather}
where $\lambda \sim \mathrm{Beta}(\alpha)$ for some hyperparameter $\alpha$ that controls the strength of interpolation. 
We use $\alpha = 0.4$ as it gives the best results within our experiments.

\paragraph{(b) Pre-processing: Massaging}
We implement massaging 
proposed by Kamiran and Calders \cite{massaging} by 
producing a modified dataset 
by relabelling the same number of instances from the favoured community with a favourable label and instances from the deprived community with an unfavourable label. 
For a dataset $D$, the favoured community ($s_\mathrm{fav}$) refers to the demographic group with a higher probability of belonging to the favourable class $y_+$, and the other group is called the deprived community ($s_\mathrm{dep}$). Since we are interested in detecting depression, $y_+$ refers to the depressed class. 
After relabelling, the distribution of classes is unchanged, but the class distribution is now the same across both genders, and the gender ratio is equal in both classes.

\paragraph{(c) In-processing: Reweighing}

We implemented reweighing which calculates weights $\beta_i$ for each example. % but the weights 
%which 
We then modify the original loss function (e.g., Cross-Entropy Loss) for a batch $B$: 
\begin{equation}\footnotesize
    \mathcal{L}(B) = \frac{1}{|B|} \sum_{(\mathbf{x}_i, y_i) \in B} \mathcal{L}_{CE}(\mathbf{x}_i, y_i), 
\end{equation}
by 
%that are 
directly incorporating $\beta_i$ into the loss function as follows: 
\begin{equation}\footnotesize
    \mathcal{L}_w(B) = \frac{1}{|B|} \sum_{(\mathbf{x}_i, y_i) \in B} \beta_i \, \mathcal{L}_{CE}(\mathbf{x}_i, y_i). 
\end{equation}
The weights are chosen according to:
\begin{equation}\footnotesize
%\footnotesize
    \beta_i = \frac{P(Y=y_i) \times P(S=z_i)}{P(Y=y_i \wedge S=z_i)} = \frac{P(Y=y_i)}{P(Y=y_i|S=z_i)},
\end{equation}
as suggested by Calders et al.\ \cite{reweighing}.

\paragraph{(d) In-processing: Regularisation}

We implement regularisation similar to \cite{reg2} where
we write $B_k$ for the instances of  group $s_k$ in  batch $B$. The True Positive Rate (TPR) for $s_k$ can be approximated as: 
\begin{equation}\footnotesize
    \mathrm{TPR}_k(B) = \frac{\sum_{(\mathbf{x}_i, y_i) \in B_k} y_i \, f(\mathbf{x}_i)}{\sum_{(\mathbf{x}_i, y_i) \in B_k} y_i},
\end{equation}
and the False Positive Rate (FPR) for the group $s_k$ is similarly defined as: 
\begin{equation}\footnotesize
    \mathrm{FPR}_k(B) = \frac{\sum_{(\mathbf{x}_i, y_i) \in B_k} (1 - y_i) f(\mathbf{x}_i)}{\sum_{(\mathbf{x}_i, y_i) \in B_k} (1 - y_i)}.
\end{equation}
%
%We define 
%
We can define the differences between TPRs and FPRs as respectively given by: 
\begin{equation}\footnotesize
    d_t(B) = |\mathrm{TPR}_0(B) - \mathrm{TPR}_1(B)|,
\end{equation}
and 
\begin{equation}\footnotesize
    d_f(B) = |\mathrm{FPR}_0(B) - \mathrm{FPR}_1(B)|.
\end{equation}
These %difference 
terms are then used to define the new loss function %which is 
\begin{equation}\footnotesize
    \mathcal{L}_\mathrm{reg}(B) = \mathcal{L}(B) + \lambda_{EOpp} \, d_t(B) + \lambda_{EOdd} \, d_f(B),
\end{equation}
where $\lambda_{EOpp}$ and $\lambda_{EOdd}$ are hyperparameters to be tuned.

\paragraph{(e) Post-processing: Reject Option Classification (ROC)}

We adopt the ROC by Kamiran et al.\ \cite{ROC} 
which attempts to improve fairness by re-classifying
the predictions that fall in a region around the decision boundary parameterised by $\tau$. 
More formally, if a sample $\x_i$ that falls in the ``critical'' region $1-\tau \leq p(y | \x_i) \leq \tau$ where $0.5 \leq \tau \leq 1$, we reclassify $\x_i$ as $y$ if $\x_i$ belongs to a minority group. Otherwise, %i.e. 
when $p(y | \x_i) > \tau$,  we accept the predicted output class $y$. 
We set $\tau=0.6$ as suggested by Kamiran et al. \cite{ROC}.

\begin{comment}

\begin{figure}
    \centering
    \includegraphics[width=\columnwidth]{Method.png}
    \emph{\caption{An illustration of how the proposed method fits into the pipeline in the second stage of training.}
    \label{fig:new_method}}
\end{figure}

\end{comment}

%%%%%%%%%%%%%%%%%%%%%%%%%%%%
%%%%%%%%%%%%%%%%%%%%%%%%%%%%

\subsection{Evaluation Measures} \label{sect: measures}

We use $s_0$ and $s_1$ to denote the minority and majority group respectively.

\vspace{-0.3cm}
\paragraph{Prediction Measures.}
We use the commonly used measures, Accuracy ($\mathcal{M}_{Acc}$), Precision ($\mathcal{M}_{P}$) 
and F1 ($\mathcal{M}_{F1}$), to evaluate prediction quality.

\vspace{-0.3cm}
\paragraph{Fairness Measures.}

We use the most prevalent metrics \cite{hort2022bias,pessach2022review,kuzucu2024uncertainty} 
and outline how each quantifies a different aspect of fairness:

\begin{itemize}   
\item\textbf{Statistical Parity} or demographic parity, is based on predicted outcome $\hat{Y}$ and independent of actual outcome $Y$:
    \begin{equation}\footnotesize
    \label{eqn:SP}
     \mathcal{M}_{SP}= \frac{P(\hat{Y}=1|s_0 ) }{ P(\hat{Y}=1 | s_1)} .
    \end{equation}
    In order for a classifier to be deemed fair, $P(\hat{Y}=1 | s_1) = P(\hat{Y}=1|s_0 ) $.

\item\textbf{Equal opportunity} states that both demographic groups $s_0$ and $s_1$ should have equal True Positive Rate (TPR). 
    \begin{equation}\footnotesize
    \label{eqn:EOpp}
    \mathcal{M}_{EOpp} = \frac{P(\hat{Y}=1|Y=1, s_0 )}{P(\hat{Y}=1 | Y=1, s_1)}.
    \end{equation}
    In order for a classifier to be deemed fair,  $P(\hat{Y}=1 | Y=1, s_1) = P(\hat{Y}=1|Y=1, s_0 ) $.
     
\item\textbf{Equalised odds} can be considered as a generalization of Equal Opportunity where the rates are not only equal for $Y=1$, but for all values of $Y \in \{1, ... k\}$, i.e.: 
    \begin{equation}\footnotesize
        \label{eqn:EOdd}
        \mathcal{M}_{EOdd} = \frac{P(\hat{Y}=1|Y=i, s_0 )}{P(\hat{Y}=1 | Y=i, s_1)} .
    \end{equation}
    In order for a classifier to be deemed fair, $P(\hat{Y}=1 | Y=i, s_1) = P(\hat{Y}=1|Y=i, s_0 ),  \forall  i \in \{1, ... k\}$.

\item\textbf{Equal Accuracy} states that both subgroups $s_0$ and $s_1$ should have equal rates of accuracy.
    \begin{equation}\footnotesize
        \label{eqn:Wacc}
        \mathcal{M}_{EAcc} = \frac{\mathcal{M}_{ACC,s_0}}{\mathcal{M}_{ACC,s_1}} .
    \end{equation}
\end{itemize}
The ideal score of 1 indicates that both measures are equal for both groups and is thus considered ``perfectly fair''. 
For practical experimental purposes, we adopt the approach of existing literature which considers $0.80$ and $1.20$ as the acceptable lower and upper fairness bounds respectively \cite{cheong2024fairrefuse}.

%%%%%%%%%%%%%%%%%%%%%%%%%%%%%%%%%%%%%%%%%%%%%%%%%%%
%%%%%%%%%%%%%%%%%%%%%%%%%%%%%%%%%%%%%%%%%%%%%%%%%%%

\section{Experimental Setup}
%\vspace{-3mm}
%\subsection{Datasets and Implementation Details}
%%%%%%%%%%%%%%%%%%%%%%%%%%%%
%%%%%%%%%%%%%%%%%%%%%%%%%%%%
\subsection{Datasets}

%\textcolor{teal}{All authors provided dataset splits which we adhered to.}

%\vspace{-2mm}
\paragraph{Mumtaz} was recorded using an EEG cap with 19 electro-gel sensors, placed according to the 10-20 system sampled at 256 Hz.
Data collection was performed during EC and EO conditions for 5 minutes. 
64 participants were recruited but the dataset only contains data from 58 individuals with 37 males and 21 females. 30 of them were diagnosed with depression based on the DSM criteria. The rest were age-matched healthy controls \cite{mumtaz_paper}.

\vspace{-2mm}
\paragraph{MODMA} consists of 
several sets of EEG data from clinically depressed patients and matching healthy controls. 
We utilise one set of EEG data collected with a 128-channel HydroCel Geodesic Sensor Net, which contains 5-minute-long EC resting-state EEG signals 
sampled at
250 Hz. 
24 out of the 53 participants were diagnosed with depression based on the DSM criteria. MODMA contains data from 33 males and 20 females so females are the minority \cite{modma}.

\vspace{-2mm}
\paragraph{Rest} contains resting-state EEG data with 64 Ag/AgCl electrodes using a SynAmps 2 system sampled at 500 Hz. 
121 participants were involved, with 46 belonging to the depressed group based on the BDI scores. Rest contains data from 47 males and 74 females so males are the minority \cite{rest_paper}.

\begin{table}[h]
    \centering
    \addtolength{\tabcolsep}{-0.5mm}
    \scalebox{0.9}
    {\begin{tabular}{l|ccc|ccc|ccc}
         & \multicolumn{3}{|c|}{\textbf{Mumtaz}} & \multicolumn{3}{|c|}{\textbf{MODMA}} & \multicolumn{3}{|c}{\textbf{Rest}} \\ \cline{2-10}
         & M & F & T & M & F & T & M & F & T \\ \hline
         Depressed & 17 & 13 & 30 & 13 & 11 & 24 & 12 & 34 & 46 \\
         Healthy & 20 & 8 & 28 & 20 & 9 & 29 & 35 & 40 & 75 \\
         Total & 37 & 21 & 58 & 33 & 20 & 53 & 47 & 74 & 121
    \end{tabular}}
    \emph{\caption{Dataset distribution. F: female. M: male. T: total. Females are the minority in Mumtaz and MODMA whereas males are the minority in Rest.}
    \label{tab:datasets}}
\end{table}

%%%%%%%%%%%%%%%%%%%%%%%%%%%%
%%%%%%%%%%%%%%%%%%%%%%%%%%%%
\subsection{ML Models}
We implement three ML models within our experimentation for comprehensiveness. The EEG data 
consists of signals arranged according to the electrode positioning which are pre-processed and fed into the ML models as outlined below.

\textbf{Deep-Asymmetry} \cite{Deep-Asymmetry} first forms a matrix to represent the pair-wise differences between the relative power of each channel for each frequency band (e.g. alpha). This matrix is provided to a CNN-based model with 3 convolutional layers. 

\textbf{GTSAN} \cite{GTSAN} first extracts the power spectral density features of bands for 1-second segments, then performs z-score standardisation. The flattened feature vectors are fed into both a GRU and a network of separable and dilated causal convolutional layers. The outputs are passed to an attention layer followed by a fully connected layer. 

\textbf{1DCNN-LSTM} \cite{1DCL} 
involves one-dimensional convolutional layers and long short-term memory (LSTM) layers
%This model has an architecture 
similar to \cite{Ay}. 

%%%%%%%%%%%%%%%%%%%%%%%%%%%%
%%%%%%%%%%%%%%%%%%%%%%%%%%%%
\subsection{Implementation Details}

\label{sec:implm_training}

All dataset owners provided dataset splits which we adhered to in our experiments.
We tuned the hyperparameters for each dataset across all the different algorithms separately.

\textbf{Deep-Asymmetry}. We use the Adam optimiser for all experiments.
For Mumtaz, the model was trained for 10 epochs at a learning rate %initialised as 
of 0.0001, mini-batch size of 75 and dropout rate of 0.25. 
For MODMA, we train the model for 50 epochs 
at a learning rate of 0.00002. We also added L2 regularisation of strength 0.01 to the dense layer. For Rest, the learning rate was 0.00005 and the number of epochs was 20. 

\textbf{GTSAN}. For MODMA, we train the model for 150 epochs with early stopping using a RMSprop optimiser with a learning rate of 0.002, mini-batch size of 128 and dropout rate of 0.2. 
For Mumtaz, the optimal settings include an RMSprop optimiser with learning rate initialised as 0.01, dropout rate of 0.2, batch size of 128, training time of 150 epochs (with early stopping)
and addition of L2 regularisation of strength 0.002 to all layers. 

\textbf{1DCNN-LSTM}. For Mumtaz, we trained the model for 20 epochs 
with a learning rate of 0.0004, mini-batch size of 50 and dropout rate of 0.5. 
For MODMA, we trained the model for 30 epochs 
with the same learning rate and mini-batch size but a dropout rate of 0.2. 
For Rest, we trained the model for 30 epochs 
with a 0.0004 learning rate, mini-batch size of 128 and dropout rate of 0.3. We also added L2 regularisation of strength 0.001 to the convolutional and LSTM layers.

%%%%%%%%%%%%%%%%%%%%%%%%%%%%%%%%%%%%%%%%%%%%%%%%%%%%%%%%%%%%%%%%%%%
%%% Results %%% 
%%%%%%%%%%%%%%%%%%%%%%%%%%%%%%%%%%%%%%%%%%%%%%%%%%%%%%%%%%%%%%%%%%%

%%%%%%%%%%%%%%%%%%%%
% Mumtaz
%%%%%%%%%%%%%%%%%%%%

\begin{table}[ht]
    %\scriptsize
    \footnotesize
    \addtolength{\tabcolsep}{-0.5mm}
    \centering
    %\captionsetup{font=scriptsize}
    %\scalebox{0.62}{
    \begin{tabular}{ll|c|ccccc}
         %& & \multicolumn{7}{|c}{\textbf{Mumtaz}} \\ \cline{3-9}
         & & Base & Aug & Mas & RW & Reg & ROC  \\ \hline
         \multirow{7}{*}{\centering \begin{sideways}\textbf{Deep-Asymmetry}\end{sideways}} & \multirow{1}{*}{\centering Accuracy} & 0.985 & \textbf{0.985} & 0.970 & 0.977 & 0.978 & \textbf{0.985} \\
 
         & \multirow{1}{*}{\centering Precision} & \textbf{0.983} & 0.981 & 0.972 & 0.974 & 0.981 & 0.980  \\
         & \multirow{1}{*}{\centering F1} & 0.985 & \textbf{0.986} & 0.971 & 0.978 & 0.978 & \textbf{0.986} \\
         \cline{2-8}

         & \multirow{1}{*}{\centering $\mathcal{M}_{SP}$} & \unfair{1.324} & \unfair{1.327} & \unfair{\textbf{1.289}} & \unfair{1.313} & \unfair{1.309} & \unfair{1.334} \\
         & \multirow{1}{*}{\centering $\mathcal{M}_{EOpp}$} & 0.995 & \textbf{1.000} & 0.981 & 0.994 & 0.991 & 1.001   \\
         & \multirow{1}{*}{\centering $\mathcal{M}_{EOdd}$} & 0.801 & \unfair{0.587} & 0.881 & \unfair{0.773} & \textbf{0.906} & 1.188 \\
         & \multirow{1}{*}{\centering $\mathcal{M}_{EAcc}$} & 1.003 & 1.007 & \textbf{0.998} & 1.003 & 1.002 & 1.004 \\
\hline
         
         \multirow{7}{*}{\centering \begin{sideways}\textbf{GTSAN}\end{sideways}} & Accuracy &  0.778 & 0.771 & 0.729 & 0.739 & 0.721 & \textbf{0.781}  \\
         & Precision & 0.781 & \textbf{0.785} & 0.769 & 0.728 & 0.743 & 0.770  \\
         & F1 & 0.811 & 0.802 & 0.757 & 0.788 & 0.758 & \textbf{0.818} \\ 
         \cline{2-8}

         & $\mathcal{M}_{SP}$ & 1.052 & 1.021 & 0.929 & \textbf{1.016} & 1.025 & 1.193 \\
         & $\mathcal{M}_{EOpp}$ & 1.032 & 0.968 & 0.933 & \textbf{1.013} & 0.917 & 1.115  \\
         & $\mathcal{M}_{EOdd}$ & \unfair{0.468} & \unfair{0.545} & \unfair{0.369} & \unfair{0.610} & \textbf{0.901} & \unfair{0.793} \\
         & $\mathcal{M}_{EAcc}$ & 1.148 & 1.076 & 1.078 & 1.168 & \textbf{0.989} & 1.160 \\ 
         \hline

         \multirow{7}{*}{\centering \begin{sideways}\textbf{1DCNN-LSTM}\end{sideways}} & Accuracy & \textbf{0.995} & 0.994 & 0.889 & 0.951 & 0.856 & 0.992  \\
         & Precision & \textbf{1.000} & 0.997 & 0.875 & 0.987 & 0.800 & 0.994   \\
         & F1 & \textbf{0.995} & 0.994 & 0.895 & 0.951 & 0.874 & 0.992  \\ 
         \cline{2-8}
         & $\mathcal{M}_{SP}$ & 1.115 & 1.124 & 0.933 & \textbf{0.959} & 0.946 & 1.133 \\
         & $\mathcal{M}_{EOpp}$ & \textbf{1.001} & 1.013 & 0.892 & 0.875 & 0.908 & \textbf{1.001} \\
         & $\mathcal{M}_{EOdd}$ & \unfair{$\infty$} & \unfair{0.000} & \unfair{0.632} & \unfair{0.000} & \unfair{\textbf{0.795}} & \unfair{$\infty$} \\
         & $\mathcal{M}_{EAcc}$ & \textbf{1.000} & 1.008 & 0.972 & 0.938 & 0.988 & 0.991  \\
    \end{tabular}%}
    \emph{\caption{Summary of performance and fairness results for the \textbf{Mumtaz} dataset. \emph{Abbreviations: Base: Baseline. Aug: Data augmentation. Mas: Massaging. RW: Reweighing. Reg: Regularisation. %Reg+: Proposed method. 
    Baseline results are the fairness results before any bias mitigation is employed.
    \textbf{Bold} values indicate the best results for each metric. 
    \textcolor{red}{Red} indicates values which fall outside the 0.80-1.20 fairness range.
    }}
    \label{tab:dataset1}
    }
    \vspace{-3mm}
\end{table}

%%%%%%%%%%%%%%%%%%%%
% MODMA
%%%%%%%%%%%%%%%%%%%%

\begin{table}[ht]
    \centering
    \addtolength{\tabcolsep}{-0.5mm}
    \footnotesize
    \begin{tabular}{ll|c|ccccc}
         & & Base & Aug & Mas & RW & Reg & ROC \\ \hline
         \multirow{7}{*}{\centering \begin{sideways}\textbf{Deep-Asymmetry}\end{sideways}} & \multirow{1}{*}{\centering Accuracy} & 0.911 & 0.913 & 0.874 & 0.873 & 0.865 & \textbf{0.917}  \\
         & \multirow{1}{*}{\centering Precision} & 0.916 & 0.922 & 0.884 & 0.884 & \textbf{0.949} & 0.924  \\
         & \multirow{1}{*}{\centering F1} & 0.901 & 0.902 & 0.856 & 0.856 & 0.833 & \textbf{0.909}  \\
         \cline{2-8}

         & \multirow{1}{*}{\centering $\mathcal{M}_{SP}$} & \unfair{1.256} & \unfair{1.244} & 1.169 & 1.147 & \textbf{1.049} & \unfair{1.357}  \\
         & \multirow{1}{*}{\centering $\mathcal{M}_{EOpp}$} & 0.899 & 0.923 & 0.821 & 0.820 & \unfair{0.750} & \textbf{0.955}  \\
         & \multirow{1}{*}{\centering $\mathcal{M}_{EOdd}$} & \unfair{1.826} & \unfair{\textbf{1.223}} & \unfair{1.791} & \unfair{1.482} & \unfair{1.472} & \unfair{2.352} \\
         & \multirow{1}{*}{\centering $\mathcal{M}_{EAcc}$} & 0.925 & \textbf{0.951} & 0.871 & 0.878 & 0.846 & 0.942 \\
\hline
         
         \multirow{7}{*}{\centering \begin{sideways}\textbf{GTSAN}\end{sideways}} & Accuracy & \textbf{0.984} & 0.981 & 0.931 & 0.846 & 0.931 & 0.947 \\

         & Precision & \textbf{1.000} & 0.990 & 0.889 & 0.972 & 0.878 & 0.925 \\
         & F1 & \textbf{0.983} & 0.980 & 0.931 & 0.813 & 0.932 & 0.945  \\ 
\cline{2-8}

         & $\mathcal{M}_{SP}$ & \unfair{1.469} & \unfair{1.433} & \unfair{1.446} & \unfair{\textbf{1.309}} & \unfair{1.454} & \unfair{1.730}  \\
         & $\mathcal{M}_{EOpp}$ & 1.062 & 1.055 & 0.948 & 0.957 & \textbf{0.985} & 1.0625  \\
         & $\mathcal{M}_{EOdd}$ & \unfair{$\infty$} & \unfair{0.000} & \unfair{4.453} & \unfair{\textbf{1.206}} & \unfair{3.216} & \unfair{$\infty$}  \\
         & $\mathcal{M}_{EAcc}$ & \textbf{1.025} & 1.031 & 0.895 & 0.927 & 0.929 & 0.919 \\ 
         \hline

         \multirow{7}{*}{\centering \begin{sideways}\textbf{1DCNN-LSTM}\end{sideways}} & Accuracy & 0.956 & \textbf{0.984} & 0.832 & 0.981 & 0.949 & 0.957  \\
         & Precision & 0.992 & \textbf{0.996} & 0.829 & 0.969 & 0.970 & 0.992  \\
         & F1 & 0.950 & \textbf{0.982} & 0.810 & 0.979 & 0.942 & 0.950 \\ 
         \cline{2-8}
         & $\mathcal{M}_{SP}$ & \unfair{1.596} & \unfair{1.449} & \textbf{1.001} & \unfair{1.374} & \unfair{1.606} & \unfair{1.596}  \\
         & $\mathcal{M}_{EOpp}$ & 1.163 & 1.063 & \unfair{0.729} & \textbf{1.006} & 1.154 & 1.163  \\
         & $\mathcal{M}_{EOdd}$ & \unfair{$\infty$} & \unfair{$\infty$} & \unfair{\textbf{1.395}} & \unfair{2.478} & \unfair{5.948} & \unfair{$\infty$}  \\
         & $\mathcal{M}_{EAcc}$ & 1.047 & 1.020 & 0.832 & \textbf{0.992} & 1.033 & 1.046  \\
    \end{tabular}%}
    \emph{\caption{Results for the \textbf{MODMA} dataset. \emph{Abbreviations and remarks: Same as those in Table \ref{tab:dataset1}.}}
    \label{tab:dataset2}
    }
%    \vspace{-3mm}
\end{table}

%%%%%%%%%%%%%%%%%%%%
% Rest
%%%%%%%%%%%%%%%%%%%%

\begin{table}[ht]
    \centering
    \footnotesize
    \addtolength{\tabcolsep}{-0.5mm}
    \begin{tabular}{ll|c|ccccc}
         & & Base & Aug & Mas & RW & Reg & ROC \\ \hline
         \multirow{7}{*}{\centering \begin{sideways}\textbf{Deep-Asymmetry}\end{sideways}} & \multirow{1}{*}{\centering Accuracy} & 0.953 & \textbf{0.953} & 0.922 & 0.927 & 0.930 & 0.951 \\
 
         & \multirow{1}{*}{\centering Precision} & 0.955 & 0.966 & 0.905 & 0.910 & 0.948 & \textbf{0.971}  \\
 
         & \multirow{1}{*}{\centering F1} & \textbf{0.933} & 0.933 & 0.891 & 0.899 & 0.899 & 0.931  \\

\cline{2-8}

         & \multirow{1}{*}{\centering $\mathcal{M}_{SP}$} & \unfair{0.539} & \unfair{0.535} & \unfair{\textbf{0.590}} & \unfair{0.586} & \unfair{0.558} & \unfair{0.561}  \\
         & \multirow{1}{*}{\centering $\mathcal{M}_{EOpp}$} & 1.025 & 1.034 & \textbf{1.021} & 1.027 & 1.029 & 1.037 \\
         & \multirow{1}{*}{\centering $\mathcal{M}_{EOdd}$} & \unfair{1.315} & \textbf{0.992} & \unfair{1.327} & \unfair{1.285} & \unfair{1.443} & \unfair{1.771} \\
         & \multirow{1}{*}{\centering $\mathcal{M}_{EAcc}$} & 1.021 & 1.028 & \textbf{1.011} & 1.012 & 1.029 & 1.017  \\

         \hline

         \multirow{7}{*}{\centering \begin{sideways}\textbf{1DCNN-LSTM}\end{sideways}} & Accuracy & 0.889 & 0.871 & 0.826 & \textbf{0.891} & 0.886 & 0.882  \\

         & Precision & \textbf{0.897} & 0.827 & 0.771 & 0.867 & 0.844 & 0.876  \\
         & F1 & 0.846 & 0.832 & 0.772 & \textbf{0.854} & 0.851 & 0.839 \\ 
         \cline{2-8}
         & $\mathcal{M}_{SP}$ & \unfair{0.593} & \unfair{0.605} & \textbf{0.951} & \unfair{0.658} & \unfair{0.612} & \unfair{0.658}  \\
         & $\mathcal{M}_{EOpp}$ & 0.978 & 1.047 & 1.148 & 1.114 & 1.012 & \textbf{0.997} \\
         & $\mathcal{M}_{EOdd}$ & \unfair{1.468} & \unfair{0.736} & \unfair{2.966} & 1.089 & \textbf{0.957} & \unfair{2.097}  \\
         & $\mathcal{M}_{EAcc}$ & 1.015 & 1.054 & 0.949 & 1.047 & 1.018 & \textbf{0.994}  \\
    \end{tabular}%}
    \emph{\caption{Results for the \textbf{Rest} dataset. \emph{Abbreviations and remarks: Identical to those in Table \ref{tab:dataset1}.}}
    \label{tab:dataset3}
    }
    \vspace{-3mm}
\end{table}

\section{Results}

%We present our results in relation to the two RQs outlined in Section \ref{sect:intro}.

\vspace{-2mm}
\paragraph{Mumtaz}
\textbf{RQ 1:} With reference to Table \ref{tab:datasets}, we see that dataset bias is present. 
Mumtaz is the most imbalanced dataset with around 76\% more males than females.
%
%In addition, 
With reference to Table \ref{tab:dataset1}, we also see that for the baseline methods, algorithmic bias is present across some fairness measures such as $\mathcal{M}_{SP}$ for Deep-Asymmetry and $\mathcal{M}_{EOdd}$ for GTSAN and 1DCNN-LSTM. 
\textbf{RQ 2:} We see that existing bias mitigation methods are not consistently effective at bias mitigation. All five methods are unable to mitigate the bias present across $\mathcal{M}_{SP}$ and $\mathcal{M}_{EOdd}$ for Deep-Asymmetry and GTSAN and 1DCNN-LSTM respectively. 

\vspace{-2mm}
\paragraph{MODMA}
\textbf{RQ 1:} With reference to Table \ref{tab:datasets}, we see that dataset bias is present within MODMA where the number of males and females has a relative difference of 65\%.
Moreover, there is a stronger presence of algorithmic bias within the baseline methods as evidenced by the $\mathcal{M}_{SP}$ and $\mathcal{M}_{EOdd}$ values across all three methods in Table \ref{tab:dataset2}.
\textbf{RQ 2:} 
Across Deep-Asymmetry, data massaging, loss reweighing and regularisation are all effective at mitigating the bias as measured using $\mathcal{M}_{SP}$ and $\mathcal{M}_{EOdd}$.
Across GTSAN and 1DCNN-LSTM, data massaging, loss reweighing and regularisation reduce the degree algorithmic bias across $\mathcal{M}_{EOdd}$ as well. 
%
%There is no bias mitigation method that consistently performs the best across all algorithms and measures.
%
% For instance, data massaging gives the fairest outcome across $M_{SP}$ whereas loss reweighing gives the fairest outcome across $M_{EOpp}$ and $M_{EAcc}$ across 1DCNN-LSTM.

\vspace{-2mm}
\paragraph{Rest}
\textbf{RQ 1:} With reference to Table \ref{tab:datasets}, we see that dataset bias is less pronounced compared to the other two datasets. The number of females is around 57\% greater than males, so males are the minority. 
With reference to Table \ref{tab:dataset3}, we see that algorithmic bias is present as measured according to  $M_{SP}$ and  $M_{EOdd}$ but not $M_{EOpp}$ and  $M_{EAcc}$.
\textbf{RQ 2:} Data augmentation provided the fairest outcome across $M_{EOdd}$ for Deep-Asymmetry.
Otherwise, we do not see evidence of effective bias mitigation across $M_{SP}$ and  $M_{EOdd}$ as both measures are consistently poor for both Deep-Asymmetry and 1DCNN-LSTM across all bias mitigation methods.
Across the other fairness measures, ROC provided the fairest outcome across $M_{EOpp}$ and $M_{EAcc}$ for 1DCNN-LSTM.

%%%%%%%%%%%%%%%%%%%%%%%%%%%%
%%%%%%%%%%%%%%%%%%%%%%%%%%%%

\section{Discussion \& Conclusion}

In this paper, we undertake the very first evaluation of ML bias for depression detection using EEG data.
We evaluate the gender fairness of three different algorithms for depression detection using EEG data %trained 
across three different datasets. 
We apply five different pre-, in- and post-processing bias mitigation techniques to evaluate the efficacy of existing bias mitigation methods.
To answer \textbf{RQ 1}, our experiments indicate that existing datasets and algorithms can be biased in favour of different genders and different fairness measures can give very different fairness outcomes. 
To answer \textbf{RQ 2}, our results indicate that existing bias mitigation methods are unable to address the bias present.
We hypothesise that this is due to the class imbalance highlighted in Table \ref{tab:datasets} which is supported by existing work 
\cite{jiaee}.
In addition, it also noteworthy that
%For each dataset, 
most models are able to
satisfy the \textit{weaker} equal opportunity $\mathcal{M}_{EOpp}$ and equal accuracy $\mathcal{M}_{EAcc}$ frameworks but usually not the \textit{stricter} equalised odds $\mathcal{M}_{EOdd}$ framework, which emphasises the need for researchers to use a wide range of fairness metric to address the high-stakes problem associated with depression detection using EEG data. 
The key takeaway is that a variety of fairness measures need to be used and further work needs to be done on identifying the best way to address the bias present.
%
%We hope our findings will assist others in developing fairer ML algorithms for depression detection using EEG data.
%
%As such, it is necessary for researchers to take dataset imbalances into account when addressing ML bias. 
%
%into these findings into 
%
%Pre- and in-processing techniques seems more effective at mitigating bias, but different techniques have different performance-fairness trade-offs. 
%
%The in-processing technique regularisation is especially promising as it has the best trade-offs in many settings. The proposed method can produce comparable or even better performance-fairness trade-offs than the original regularisation approach.
%
%To conclude, we show that the bias of ML algorithms for EEG-based depression detection can usually be effectively mitigated at a low or no cost of performance.

%\end{comment}

%\input{content/Angus_thesis}

%\vfill
%\pagebreak

%\clearpage
\vspace{-2mm}
\small
\paragraph{Compliance with Ethical Standards:} 
This research study was conducted retrospectively using human subject data made available in open access. Ethical approval was not required as confirmed by the license attached with the open access data.

\vspace{-2mm}
\subsection*{Acknowledgements} 
%\vspace{-2mm}
\noindent\textbf{Funding:} 
J. Cheong is supported by the Alan Turing Institute doctoral studentship, the Leverhulme Trust, and partially by the EPSRC/UKRI project ARoEq under grant ref. EP/R030782/1, and acknowledges resource support from METU. H. Gunes’ work is supported by the EPSRC/UKRI project ARoEq under grant ref. EP/R030782/1.
\noindent\textbf{Open access:} 
A Creative Commons Attribution (CC BY) licence to any Author Accepted Manuscript version arising.
\noindent\textbf{Data access:} This study involved secondary analyses of existing datasets that have been described and cited accordingly.

\small
\bibliographystyle{IEEEbib}
\bibliography{Angus,Jiaee}

\end{document}